\documentclass[10pt, a4paper]{article}
\usepackage{lrec-coling2024}
\usepackage{times}
\usepackage{url}
\usepackage{latexsym}
\usepackage{amsmath}
\usepackage{xcolor}
\usepackage{booktabs}
\usepackage{caption}
\usepackage{mathtools}%
\usepackage{usebib}
\usepackage{subcaption}

\newcommand{\model}[0]{SentEmoContext~}

\usepackage{graphicx}
\graphicspath{ {./figures/} }
\usepackage{color}
\usepackage{verbatim}
\definecolor{deepblue}{rgb}{0,0,0.5}
\definecolor{deepred}{rgb}{0.6,0,0}
\definecolor{deepgreen}{rgb}{0,0.5,0}
\definecolor{gael}{RGB}{204, 0, 255}

\title{Context-Aware Siamese Networks for Efficient Emotion Recognition in Conversation}

\name{Barbara Gendron$^{1,2}$\quad Gaël Guibon$^1$}

\address{ \\
         $^1$ LORIA, Université de Lorraine, CNRS \\
         $^2$ Université du Luxembourg \\
         \{firstname.lastname\}@loria.fr\\}

\abstract{The advent of deep learning models has made a considerable contribution to the achievement of Emotion Recognition in Conversation (ERC). However, this task still remains an important challenge due to the plurality and subjectivity of human emotions. Previous work on ERC provides predictive models using mostly graph-based conversation representations. In this work, we propose a way to model the conversational context that we incorporate into a metric learning training strategy, with a two-step process. This allows us to perform ERC in a flexible classification scenario and to end up with a lightweight yet efficient model. Using metric learning through a Siamese Network architecture, we achieve 57.71 in macro F1 score for emotion classification in conversation on DailyDialog dataset, which outperforms the related work. This state-of-the-art result is promising regarding the use of metric learning for emotion recognition, yet perfectible compared to the microF1 score obtained.
 \\ \newline  }

\begin{document}
\maketitleabstract

\section{Introduction}
\label{sec:intro}

Computer Mediated Communication (CMC) is constantly evolving and new means of communicating are emerging. With the advent of conversational agents, there is a need to detect emotions within a conversation. Although many modalities are now considered in the communication process, the textual modality still remains essential for fast and easy everyday communication, through messaging applications, social media, and other networking platforms. Textual modality, however, is ambiguous, it does not preserve the extra-linguistic context, especially for dyadic human-to-human conversations. One main ambiguity that arises in CMC is the emotional state of the speaker, often misinterpreted by humans through short, and unpolished messages. This motivates Emotion Recognition in Conversation (ERC), a trending research topic dedicated not only to identifying emotion in messages, but also on taking into account the conversational context to recognize emotions. ERC has been shown to be challenging, especially with respect to the way to represent the context~\cite{ghosal-etal-2021-exploring}. Lately, it has seen a surge of multimodal models and graph-related approaches which often try to map the pattern of each speaker and better represent the conversational context, often resulting in good performance at the cost of efficiency. One additional issue ERC models are facing is their dependency on labels, models are mainly supervised and face the issue of extreme label imbalance due to emotional utterances being so scarce. 

In this paper, we tackle these two challenges by incorporating the conversational context into metric learning, while heavily controlling the data imbalance by multiple means. Considering that we want to tackle information across emotions to make our model usable for variant of emotions that goes beyond the scope of the 6 basic emotions, we do not use supervised contrastive learning~\cite{khosla2020supervised} in our method. Instead, we focus on a two-step process to update the model both using direct label predictions through a Cross entropy loss, and relative label assignment through the contrastive loss. This two-step process is quite straight forward while using isolated elements, such as isolated utterances. However, as far as we know, the contextual representation through contrastive learning for ERC has yet to be used. This represents our main contribution in this paper as we present a model that can achieve competitive performance compared to the state of the art while rendering the adaptation to other emotion labels feasible. Thus, our model can be applied and adapted in multiple contexts requiring emotion recognition of different label granularities.

Our main contribution lies in the development of a metric-learning training strategy for emotion recognition on utterances using the conversational context. The presented model leverages sentence embeddings and Transformer encoder layers~\cite{Vaswani:2017,devlin-etal-2019-bert} to represent dialogue utterances and deploy attention on the conversational context. Our method involves Siamese Networks~\cite{Koch:2015} in the setup but can be adapted to any metric-learning model. We further demonstrate that our approach outperforms some of the latest state-of-the-art Large Language Models (LLMs) such as light versions of Falcon or LLaMA 2~\cite{touvron2023llama}. In addition, our method is efficient in the sense that it involves lightweight, adaptable and quickly trainable models, which still yield state-of-the-art performance on DailyDialog in macroF1 score with 57.71\% and satisfactory results on microF1 with 57.75\%.

In the following sections, we first review related work on ERC (Section~\ref{sec:sota}). We then dive in our methodology (Section~\ref{sec:method}) and describe the experimental setup we use (Section~\ref{sec:expe}). We then evaluate our models compared to a baseline without any conversational context and to SotA models for ERC in Section~\ref{sec:res}. Finally, we end up with key findings and perspectives for future work in section~\ref{sec:limitations}. 

We will make our code and models available on github and Hugging Face models hub.

\section{Related Work}
\label{sec:sota}

\paragraph{ERC.} Although most of the studies on ERC has been held on multi-modal datasets~\cite{sota-meld,sota-iemocap,presque-sota-iemocap}, thus leveraging multi-modality, there are still some models developed for emotion recognition on textual conversation only, whether it be on multi-modal datasets restricted to text such as IEMOCAP~\citeplanguageresource{IEMOCAP:2008} or MELD~\citeplanguageresource{MELD:2019}, or on fully textual dataset such as DailyDialog~\citeplanguageresource{Li:2017}. The advent of deep learning enables significant progress in ERC on text, starting by the use of Recurrent Neural Networks (RNN)~\cite{rumelhart:1985,jordan:1986} by~\newcite{poria-etal-2017-context}. Further work using recurring structures followed, such as DialogueRNN~\cite{majumder:2018,ghosal-etal-2020-cosmic}. This model leverages the attention mechanism~\cite{Bahdanau:2014} encountered in Transformer architecture~\cite{Vaswani:2017}. Graph-based methods also proved efficient as shown in~\cite{ghosal-etal-2019-dialoguegcn}, not only as such but also when considering external knowledge, as~\newcite{lee-choi-2021-graph} use a graph convolutional network (GCN) to perform ERC by extracting relations between dialogue instances. 

Existing work on ERC relies mainly on evaluating their model using micro F1 score excluding the majority neutral label. However, recent work actually skipped this evaluation to instead only focus on the macro version of this metric~\cite{pereira-etal-2023-context}, while other considered the Matthew Coefficient Correlation as an indication suitable for this task~\cite{guibon:2021:protoseq}.

In this work, we focus on DailyDialog, which consists in artificially human-generated conversations about daily life concerns, with utterance-wise emotion labelling. \citet{liang-etal-2022-page} propose a model based on Graph Neural Networks (GNN) and CRF that achieves 64.01\% in micro F1.

Although it is known not to provide the best performance compared to few-shot learning approaches~\cite{dumoulin:2021}, meta-learning allows better generalization through a more robust training~\cite{Finn:2017,Antoniou:2019}, which is particularly adapted in the case of emotion detection due to both variability and complexity of human feelings~\cite{Plutchik:2001}.

\paragraph{Metric learning.} As reviewed by~\cite{Hospedales:2022}, a meta-learning approach consists in a \textit{meta-optimizer} that describes meta-learner updates, a \textit{meta-representation} that stores the acquired knowledge and the \textit{meta-objective} oriented towards the desired task. This \mbox{optimization-based} meta-learning setup provides end-to-end algorithms often based on episodic scenarios~\cite{Ravi:2016,Finn:2017,Mishra:2018} that reflect the "learning to learn" strategy. Besides, learning to learn implies second order gradient computations which is costly. Palliative solutions to this problem, such as implicit differentiation~\cite{Lorraine:2020}, still involve a trade-off between performance and memory cost~\cite{Hospedales:2022}. Therefore, variants has emerged such as \textit{metric learning}, which meta-objective is the meta-representation learning itself. Starting with Siamese Networks~\cite{Koch:2015}, this model structure leverages parameter sharing between identical sub-networks to learn a distance between data samples. Relation Networks~\cite{Sung:2018} also consider a distance metric, departing from the traditional Euclidean approach. Matching Networks~\cite{Vinyals:2017} leverage training examples to identify weighted nearest neighbors. Prototypical Networks~\cite{Snell:2017} compute average class representations and utilize cosine distance for element comparison. This model has been adapted to perform ERC in a few-shot setting by~\cite{guibon:2021:protoseq} in a way that outperformed few-shot learning baselines.

In this work, we focus on Siamese Network architecture. It has the advantage to be conceptually simple, rendering it easily controllable and scalable. Nevertheless, the model structure proposed in this paper is easily adaptable to more complex meta-learning setups. Siamese Networks have been used, for instance, in NLP for intention detection on text~\cite{Ren:2020}, in computer vision for facial recognition~\cite{hayale:2023}, and in complex representation learning~\cite{Jin:2021}.

\section{Methodology}
\label{sec:method}

In this work, we use a metric-learning architecture based to learn emotions as they relate to each other, thus extracting meta-information from the data. The model is a Siamese network~\cite{Koch:2015} with three identical sub-networks, whose outputs are compared using the triplet loss~\cite{tripletInitial}. Initially applied to computer vision problems~\cite{tripletCVApplication,faceNet}, triplet loss is defined on a triplet of data samples $(a, p, n)$ so that if $a$ and $p$ belong to the same class and $n$ belongs to a different class, then: 
\begin{equation*}
\mathcal{L}(a, p, n) = \text{max}\left\{ d(a, p) - d(a, n) + \text{margin}, 0\right\}
\end{equation*}
where the margin parameter is a strictly positive number. 

\begin{figure}[h!]
    \centering
    \includegraphics[width=0.48\textwidth]{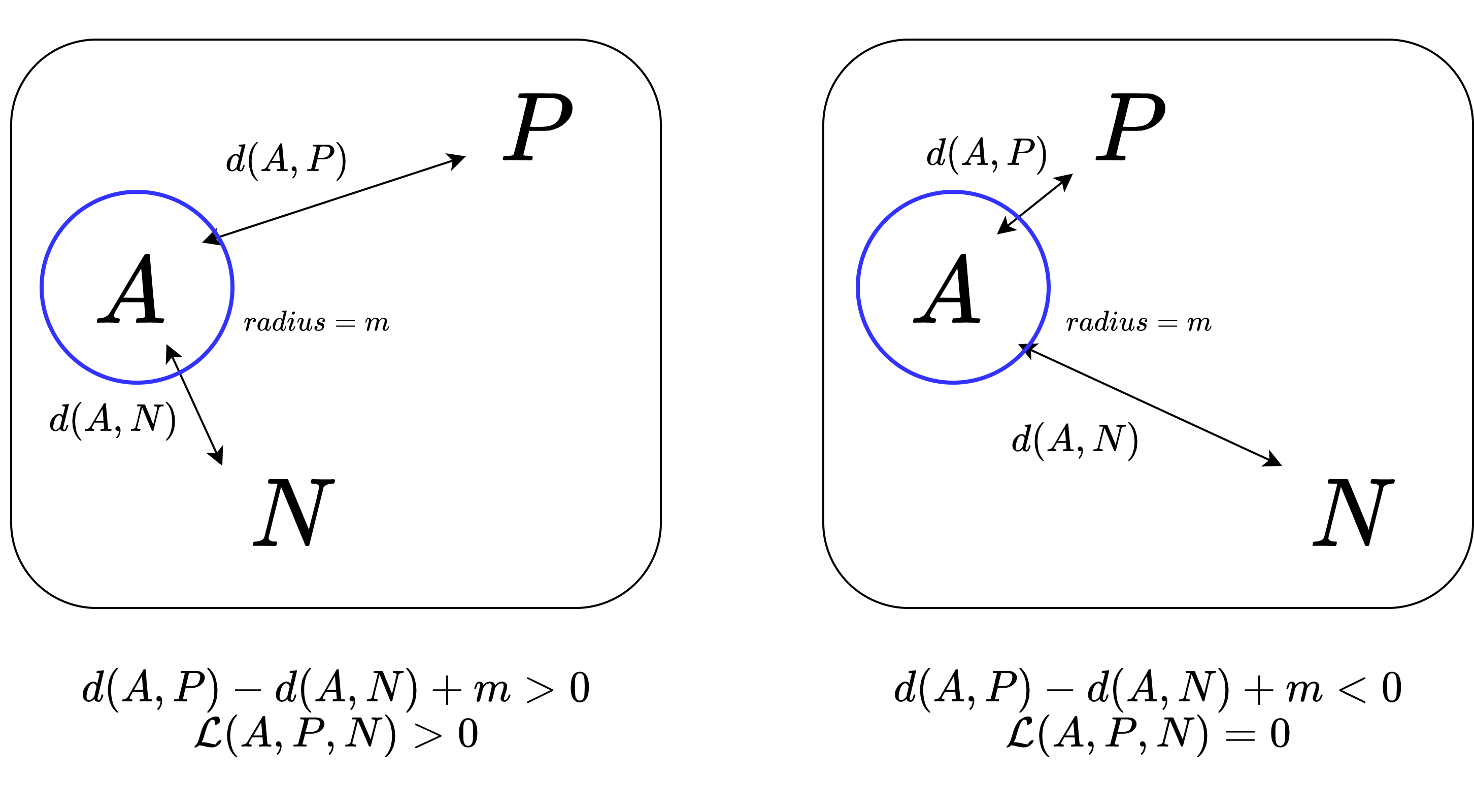}
    \caption{\centering Illustration of the triplet loss principle. Given a triplet $(A. P. N)$ corresponding to respectively \textit{anchor}. \textit{positive} and \textit{negative}. the positive sample should be closer to the anchor than the negative sample in order to minimize the triplet loss.}
    \label{fig:triplet-loss}
\end{figure}

While the triplet loss could be used in several strategies, ranging from only retrieving the most difficult triplets (when the positive is far from the anchor, meanwhile the anchor is close to the negative) to skipping the most easy ones (i.e. when the positive is closer to the anchor), we only tackle the overall strategy by considering each triplets in our data, due to the limited size of the data.

\paragraph{Isolated representations.} As the aim of our experiments is to characterize the contribution of conversational context to emotion prediction, we first developed a baseline model on isolated utterances. This formally refers to computing emotion predictions for utterances independently of their context. To do this, we first consider a mapping for each utterance word to its associated FastText embedding~\cite{bojanowski-etal-2017-enriching}. From such embeddings, aforementioned $(a, p, n)$ triplets are randomly sampled and given as input for the Siamese network, whose sub-network gradually improves in emotion prediction as triplet loss backpropagates. 

\begin{figure*}
    \centering
    \includegraphics[width=\textwidth]{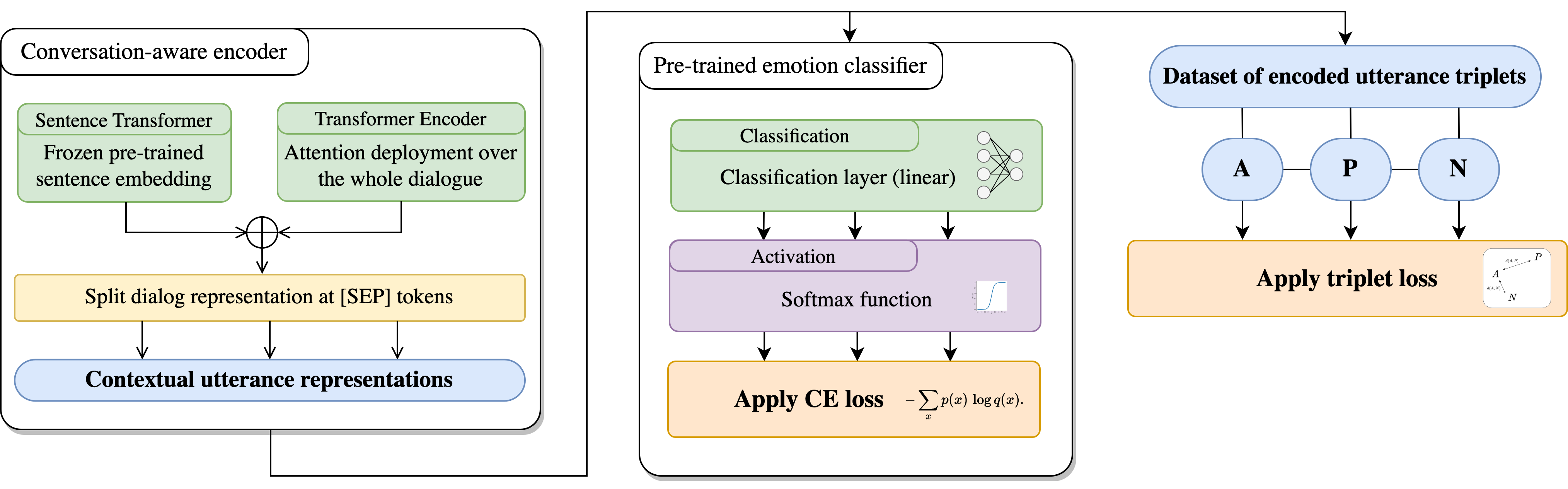}
    \caption{\centering Illustration of the three main steps of the training procedure in the case of conversation-aware emotion predictions. Both losses (CE and triplet) backpropagate in order to gradually improve the encoder.}
    \label{fig:contextual-training}
\end{figure*}

\paragraph{Contextual representations.} Regarding the contextual case, we build contextual utterance representations upon a BERT-like encoding. Sentence embeddings are preferred to word-piece embeddings (like BERT produces) as they provide lighter utterance representations. After the dialog is mapped to its associated series of pretrained embeddings, these outputs are concatenated forming a dialog representation, and contextual information is considered by deploying attention over it. Concretely, a Transformer encoder layer is stacked to the gathered frozen pre-trained embeddings. This newly conversation-aware dialog representation is then split at \texttt{[SEP]} tokens to end up with contextual representations at the utterance level, on which the emotion prediction is performed. In order to fit contextual utterance representations to the emotion prediction objective, we add an emotion classifier that is pre-trained on DailyDialog training set. The classifier is not frozen to ensure a complete backpropagation. Meanwhile, contextual representations are optimized according to the metric learning objective, using triplet loss. The whole is illustrated in figure~\ref{fig:contextual-training}. This training scenario enables both individual and relative emotion learning, in such a way that each learning phase strengthens the other. Thanks to this meta-learning setting, meta-information about emotions is extracted, and we can expect that this model is able to achieve relevant classification on unseen labels in a few-shot setting.

\section{Experimental Protocol}
\label{sec:expe}

\paragraph{Data.} All the experiments have been carried out on DailyDialog dataset~\citeplanguageresource{Li:2017} that provides more than 10,000 dialogues about daily concerns along with utterance-wise emotion labelling. In addition to providing utterance-level emotion labeling, an advantage in using DailyDialog is that it is relatively small, therefore it is quite easy to handle the entries and run tests on it. There exist six emotional labels (anger, disgust, fear, happiness, sadness and surprise) and a neutral label. Regarding emotion prediction, the evaluation is carried out only on the emotional labels following previous work procedure~\cite{ghosal-etal-2021-exploring,zhong-etal-2019-knowledge}. We use the original dataset splits (train, validation and test) from~\citeplanguageresource{Li:2017}. The main characteristics from DailyDialog dataset are visible in Table~\ref{tab:datacounts}.

\begin{table}[htbp]
    \centering
    \begin{tabular}{c|c|}
        \multicolumn{2}{c}{Daily Dialog Stats}  \\
        \toprule
        Language & English \\
        Max Msg/Conv & 35 \\
        Avg Msg/Conv & 8 \\
        Labels & 7  \\
        Emotion Labels & 6 \\
        Nb. Conv. & 13,118 \\
    \end{tabular}
    \caption{\centering Main statistics for DailyDialog dataset}
    \label{tab:datacounts}
\end{table}

\paragraph{Model specificities.} For the isolated utterance model, we consider two different types of sub-networks being simple linear layers and Long Short-Term Memory layers (LSTM)~\cite{hochreiter:1997}. In the contextual case, the sub-network is a Transformer encoder fed with sentence embeddings. We carried out experiments with three different models of pre-trained sentence Transformers available in the Python library \texttt{sentence transformers}\footnote{\url{https://www.sbert.net/}}: MPNet~\cite{Song:2020MPNet}, MiniLM~\cite{Wang:2020MiniLM} and RoBERTa~\cite{liu2019roberta}. In order to ensure a good balance, the $(a,p,n)$ triplets are made at this stage, meaning right before applying the pretrained emotion classifier, which is composed of a linear layer stacked upon one Transformer encoder layer.

\paragraph{Training specificities.} Whether it be for the isolated utterance model or for the contextual one, the emotion prediction is always performed at the utterance level, therefore the triplets are always utterance triplets. This involves balance issues as DailyDialog dataset is very imbalanced regarding emotion labels (Figure~\ref{fig:dd-distrib}). Indeed, the class rebalancing induced by sampling triplets according to a uniform distribution does not sufficiently mitigate bias during training and prevents the loss from converging due to excessive oversampling in frequent classes. Thus, we addressed the imbalance problem all along the training pipeline, by implementing a random sampler weighted with inverse label frequencies to account for the rareness of some emotional labels like \texttt{fear} or \texttt{disgust}.

\paragraph{Evaluation.} For quantitative evaluation we needed to account for both performance and relevancy of the training procedure so that generalization abilities enabled by the meta-learning architecture are actually usable. This way, we selected, in addition to usual performance metrics, a highly demanding metric: Matthews Correlation Coefficient (MCC)~\cite{Cramer+1946}. This measures a Pearson correlation~\cite{pearson:1895} between the predicted and the actual class, giving more precise information on classification quality than F1 score~\cite{Baldi:2000}. Using $TP$. $TN$. $FP$ and $FN$ as respectively the number of true positives, true negatives, false positives and false negatives, MCC was originally defined in~\cite{Matthews:1975} as:
\begin{equation}
    \text{MCC} = \frac{\mathit{TP}/ N-S \times P}{\sqrt{P S(1-S)(1-P)}}
\end{equation}

\begin{figure}[!ht]
     \centering
     \begin{subfigure}[b]{0.47\textwidth}
         \centering
         \includegraphics[width=\textwidth]{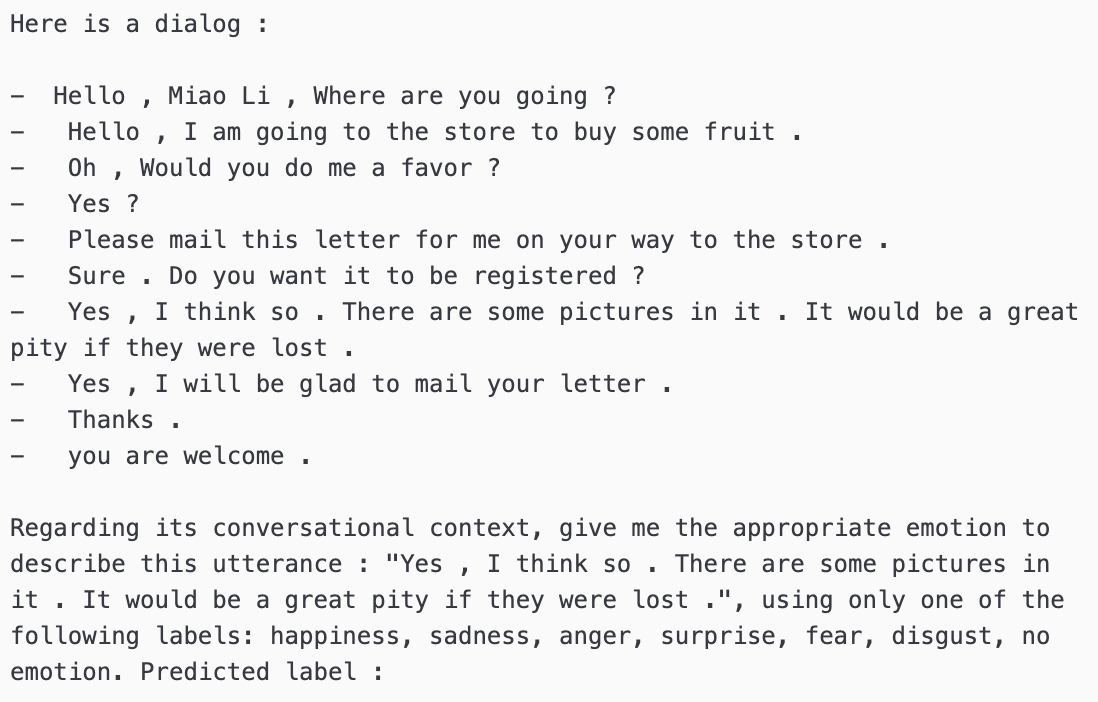}
         \caption{\centering Prompt for llama}
     \end{subfigure}
     \hfill
     \begin{subfigure}[b]{0.47\textwidth}
         \centering
         \includegraphics[width=\textwidth]{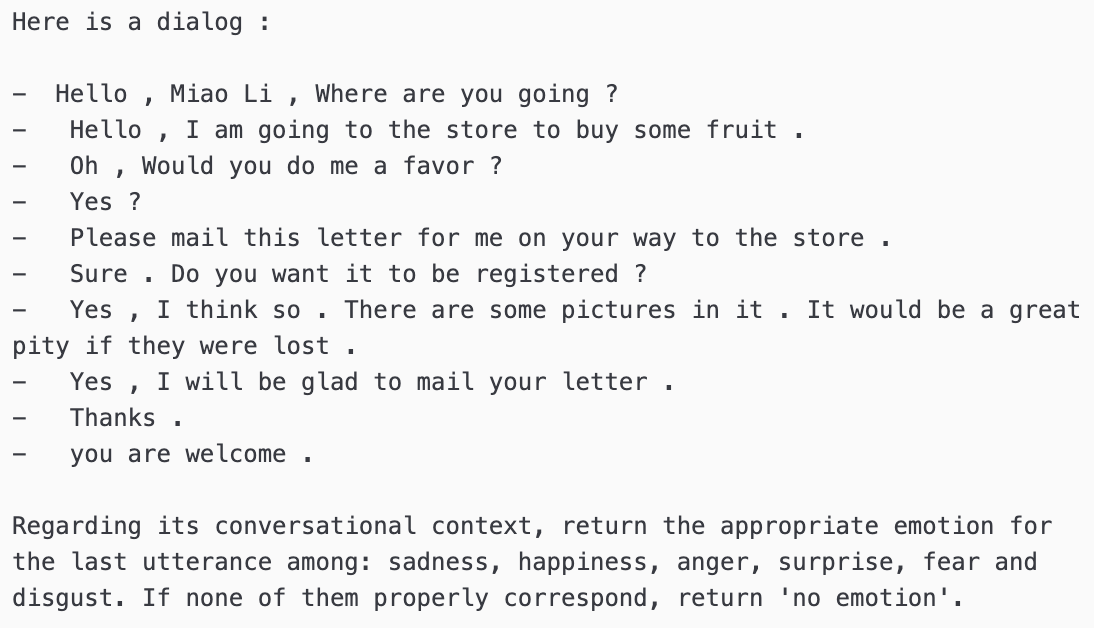}
         \caption{\centering Prompt for Falcon}
     \end{subfigure}
     \caption{\centering Prompts for llama and falcon}
     \label{fig:prompts}
\end{figure}

\paragraph{Comparison with LLMs.} In order to place the results of our isolated and contextual models into perspective, we compare our models with state-of-the-art LLMs, namely LlaMA~\cite{touvron2023llama} and Falcon~\cite{penedo2023refinedweb}. Both are considered with instruction fine-tuning and evaluated on text generation inference in a zero-shot setting. We developed a prompt asking for prediction on the last utterance of each DailyDialog test set dialog, regarding the conversational context. For both LLM, we went through an iterative process to find the most adapted prompt in the sense that the model actually generates only one label. The prompt is the same for each model of the same type (either Llama or Falcon). We experienced more difficulty on prompt tuning with Falcon as the model generates \texttt{happiness} on 86\% of DailyDialog test set. Both prompts full texts are provided in Figure~\ref{fig:prompts}.

\section{Results}
\label{sec:res}

\begin{table*}[!ht]
\centering
\begin{tabular}{@{}llll@{}}
\toprule
\textbf{Model name}                           &\textbf{ macroF1*} & \textbf{microF1*} & \textbf{MCC} \\ \midrule
\multicolumn{4}{c}{State-of-the-art models on ERC}\\ \hline
CNN+cLSTM~\cite{poria-etal-2017-context}        & --                    & 50.24             & -- \\
KET~\cite{zhong-etal-2019-knowledge}            & --                    & 53.37             & -- \\
COSMIC~\cite{ghosal-etal-2020-cosmic}           & 51.05                 & 58.48             & -- \\
RoBERTa~\cite{ghosal-etal-2020-cosmic}          & 48.20                 & 55.16             & -- \\
Rpe-RGAT~\cite{ishiwatari-etal-2020-relation}   & --                    & 54.31             & -- \\
Glove-DRNN~\cite{ghosal-etal-2021-exploring}    & 41.8                  & 55.95             & -- \\
roBERTa-DRNN~\cite{ghosal-etal-2021-exploring}  & 49.65                 & 57.32             & -- \\
CNN~\cite{ghosal-etal-2021-exploring}           & 36.87                 & 50.32             & -- \\
DAG-ERC~\cite{shen-etal-2021-directed}          & --                    & 59.33             & -- \\
TODKAT~\cite{zhu-etal-2021-topic}               &  \underline{52.56}    & 58.47             & -- \\
SKAIG~\cite{li-etal-2021-past-present}          & 51.95                 & 59.75             & -- \\
Sentic GAT~\cite{tu:2022}                       & --                    & 54.45             & -- \\
CauAIN~\cite{ijcai2022p628}                     & --                    & 58.21             & -- \\
DialogueRole~\cite{Ong_Su_Chen_2022}            & --                    & 60.95             & -- \\
S+PAGE~\cite{liang-etal-2022-page}              & --                    & \textbf{64.07}    & -- \\
DualGAT~\cite{zhang-etal-2023-dualgats}         & --                    & \underline{61.84} & -- \\
CD-ERC~\cite{pereira-etal-2023-context}         & 51.23                 & --                & -- \\
Llama2-7b~\cite{touvron2023llama}               &   9.70                &  24.92            & 0.08\\
Llama2-13b~\cite{touvron2023llama}              &  22.26                &  43.37            & 0.15\\
Falcon-7b~\cite{penedo2023refinedweb}           &  07.54                &  42.75            & 0.01\\ 
\hline
\multicolumn{4}{c}{Ours}\\ \hline
\model                                          &  \textbf{57.71}       &  57.75            & \textbf{0.49}\\
\bottomrule
\end{tabular}
\caption{\centering All results for ERC on DailyDialog. Metrics are all computed on the official test set. DRNN stands for DialogueRNN as it is called in the original paper. MCC = Matthew Coefficient Correlation. The * indicates metrics that do not include the neutral label.}
\label{tab:res}
\end{table*}

Table~\ref{tab:res} gives an overview of the different results obtained by the research community on ERC with DailyDialog. This actually shows a slow progression since 2017 where \citet{poria-etal-2017-context} proposed to evaluate the model on the micro F1 score excluding the majority class (i.e. the neutral class). This became the first baseline for this task, achieving 50.24 in micro F1 score. On the other hand, the current SotA model now achieves 64.07 in micro F1 score~\cite{liang-etal-2022-page} which amounts to a ~14 points improvement during 6 years. As visible in Table~\ref{tab:res}, the community mainly followed this pattern and evaluation scheme. However, in this paper we think it is important to also consider the macro F1 score, excluding the majority class, as it shows the overall performance on all emotions. Some work already decided to do so since 2020~\cite{ghosal-etal-2020-cosmic}, leading to an improvement of \~2.5 points in 3 years. This reinforces the claim that the ERC task is indeed challenging.

Compared to these results, our \model model achieves 57.75 in micro F1 score, which is a decent but somewhat modest result, in terms of metric comparison. However, Table~\ref{tab:res} also shows the average performance of our model over 10 runs. Our \model is SotA on the macro F1 score with 57.71 points, outperforming CD-ERC~\cite{pereira-etal-2023-context} by 6.48 points, which is considerable since they only focused on this metric, and TODKAT~\cite{zhu-etal-2021-topic} by 5.15 points. We also evaluate our model using the multiclass MCC~\cite{Matthews:1975,Baldi:2000} score in order to ensure the model is not deciding randomly. Given a MCC score ranges from -1 to 1, and 0 indicating randomness, the 0.49 MCC score of \model model indicates our approach is both balanced and accurate in terms of predictions\cite{chicco2020advantages}. Of course, we cannot compare to other ERC works with the MCC metric, as they did not used it. However, we think it is important to consider it as an additional metric to indicate the quality of the classification, minimizing the effect of the highly imbalanced data from conversations.

Given these results, our \model performs really well considering we only need \~20 minutes per epoch, and train it using only 5 epochs. This makes a striking difference with existing approaches using multiple streams per speaker~\cite{pereira-etal-2023-context}, graph modeling for context and knowledge representation~\cite{zhong-etal-2019-knowledge,li-etal-2021-past-present}, or other heavy representation in their model~\cite{liang-etal-2022-page}. In addition to this, our model is stable with a standard deviation of only 0.01 on average across the three metrics, which reinforces the quality of such an efficient approach.

\subsection{Comparison with Emotion Classifiers on Utterance Level}

Table~\ref{tab:emoclf} shows the results of direct emotion classification on utterances. For this task, we only considered the 6 emotion labels, excluding the neutral one not only from the evaluation, but also from the training. By doing so we want to determine the difference between our approach and a dedicated emotion classifier. This also serve as an ablation study for our \model model since this step is part of its training. With Table~\ref{tab:emoclf}, we can see our model leverages both the embedded conversational context and the metric learning scheme to increase all metrics. We can especially note the difference in terms of macroF1 score, which shows the importance of the triplet loss representation in our model. Indeed, the emotion utterance classifiers are trained using batches balanced on the whole training set distribution and a weighted cross entropy loss. Results shows it is not enough to deal with an extreme imbalanced data such as conversations.

\begin{table*}[!ht]
\centering
\begin{tabular}{llll}
\toprule
\textbf{Model name}                          &\textbf{ macroF1} & \textbf{microF1} & \textbf{MCC} \\ \midrule
\multicolumn{4}{c}{Pre-trained emotion utterance classifiers}\\ \hline
all-MiniLM-L6-v2         & 20.22 & 33.11  & 0.40 \\
all-mpnet-base-v2        & 14.43 & 32.90  & 0.37 \\ \hline
\multicolumn{4}{c}{Ours}\\ \hline
\model                &  \textbf{57.71}   &  \textbf{57.75}   &  \textbf{0.49}   \\
\bottomrule
\end{tabular}
\caption{Comparison with a direct emotion classification at the utterance level. The all-MiniLM-L6-v2 fine-tuning is also part of the whole \model approach.}
\label{tab:emoclf}
\end{table*}

\subsection{LLM-related Limitations}

\begin{table*}[!ht]
\centering
\begin{tabular}{@{}lllllll@{}}
\toprule
\textbf{Model name}                          & \textbf{P} & \textbf{R} &\textbf{ macroF1*} & \textbf{microF1*} & \textbf{MCC} \\ \midrule
\multicolumn{6}{c}{LLMs}\\ \hline
llama2-7b                            &  26.77  &  24.77  &   9.70  &  24.92     &  0.08   \\
llama2-13b                           &   32.63  &  83.49  &  22.26   &   43.37  &  0.15   \\
falcon-7b                           &  --  &  --   &  07.54   &  42.75  &  0.01 \\ 
\bottomrule
\end{tabular}
\caption{Results using two open-source LLMs with specific prompts. (An example of the prompt is given in Figure~\ref{fig:prompts}.}
\label{tab:llm}
\end{table*}

LLMs results on a zero-shot setting are visible in Table~\ref{tab:llm}. These serves as an indication on the performance of such models, albeit in their lightweight version, in the ERC task. Even though these generative models are not designed for this quite peculiar task, they still manage to outperform utterance emotion classifiers from Table~\ref{tab:emoclf}, which can be considered as a display of emergent capacities from LLMs~\cite{srivastava2022beyond}.

\subsection{Imbalance Factor}
\begin{figure}[htbp]
    \centering
    \includegraphics[width=0.48\textwidth]{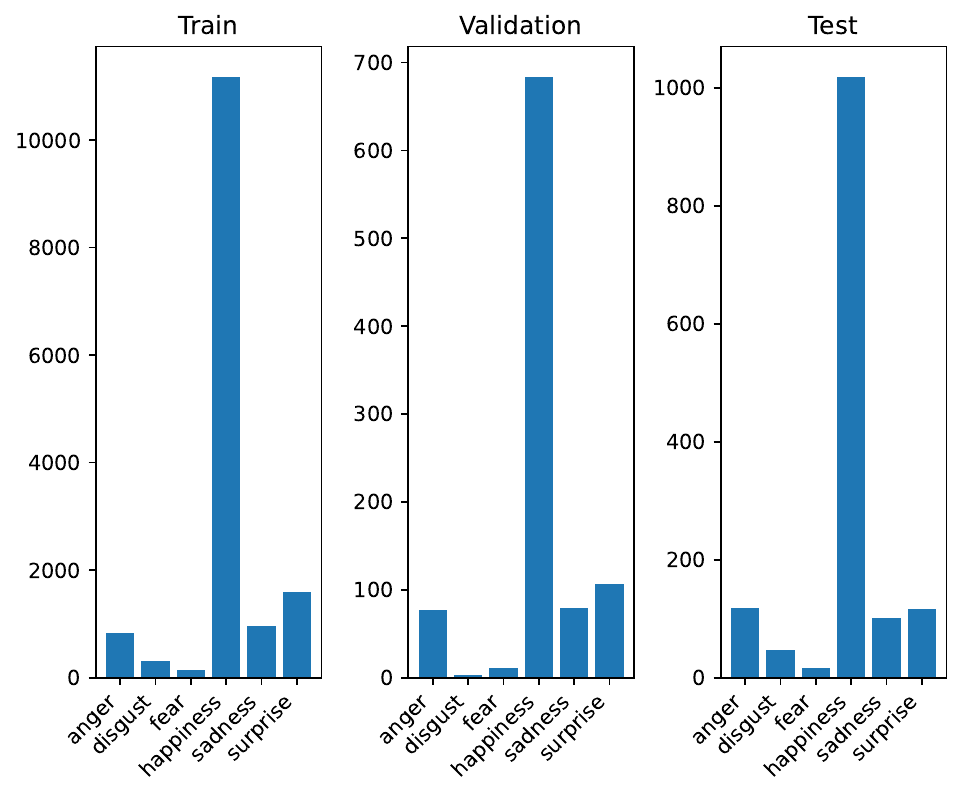}
    \caption{\centering Histograms of only the emotion label distribution in DailyDialog subsets.}
    \label{fig:dd-distrib}
\end{figure}

While Table~\ref{tab:datacounts} shows the characteristics from the dataset, it omits to present the main characteristic from conversational data in terms of emotion labels: the extreme imbalance. Most of the difficulty from ERC comes from the label definition, the context but also from the imbalance factor that prevents the model from easily learning emotion representation in the context. Figure~\ref{fig:dd-distrib} shows the distribution of the labels in DailyDialog, without the neutral one. Considering the latter is the majority label and is excluded from the evaluation metrics by all the ERC community, the fact that even in the emotion labels the data is that imbalanced proves to be challenging and needs to be addressed. We actually stem from \citet{Guibon_Labeau_Lefeuvre_Clavel_2023} to tackle the imbalance characteristic in two-steps. First, we balance the data loader to yield somewhat balanced batches given the training set weights. Second, we weight the cross entropy loss from the emotion classifier considering the remaining imbalance on each batch. 

In addition to this, in this paper we add another way to address the imbalance. By considering triplets we remove the imbalance factor while using hidden states that come from balanced representation. We think this partly explains the effectiveness and the efficiency of our model, considering its limited size compared to the related work. 

\section{Discussions \& Limitations}
\label{sec:limitations}

The work we present in this paper still possesses some limitations. We hereby draw some conclusion from them.

\subsection{LLMs Limitations}
The first limitation we faced with LLMs is the requirement of high memory GPUs to test them. This explains why in Table~\ref{tab:llm} we only consider the lightweight version of these two open source LLM. While Llama 7b and 13b gave answers in a good format, i.e. with only one label chosen, Falcon did not behave the way we wanted. In order to solve this, we look for the first mentioned emotion in the output to consider it as a label. 

Also, it is important to note that we did not want to tackle OpenAI's ChatGPT due to the fact that we do not have a clear control on the model version, size and approach used behind its API, but also because we wanted to consider open source models, and open source data as we will release both our models and source code to the community.

An additional possible limitation on LLMs is the context size. In ERC, context size is key but with LLMs adding examples in the prompt to do few-shot learning would take a lot of space in the overall context, the prompt being part of the context. This explains our decision to only consider zero-shot in this paper for LLMs, even though we should also consider prompt tuning to enhance them on this specific task.

\subsection{Model Size and Efficiency}

Our \model is efficient. It yields state-of-the-art results on macro F1 score and good results on microF1. But our model trains relatively fast and does not require a lot of epochs to converge. We think this efficiency along with the limited memory needed to train, is due to both our two-step backpropagation and to the fact that we are using utterance embedded representations with sentence transformers. Thus, our model can efficiently tackle long conversational contexts with a limited cost in memory.

\begin{table*}[!ht]
\begin{tabular}{@{}lllllll@{}}
\toprule
\textbf{Model name} & \textbf{Seq. Length} & \textbf{Tokens} & \textbf{Dimensions} & \textbf{Size} & \textbf{Parameters} & \textbf{Tr. Layers} \\ \midrule
\multicolumn{7}{c}{Pre-trained sentence transformers} \\\midrule
all-MiniLM-L6-v2  & 256  & 1bn+ & 384   & 80 MB  & 22M  & 6  \\
all-mpnet-base-v2 & 384  & 1bn+ & 768   & 420 MB & 110M & 12 \\
\midrule
\multicolumn{7}{c}{State-of-the-art LLMs}  \\\midrule
Llama-2-7b-chat-hf & 4096 & 2T & 11008 & 13 GB  & 7B   & 32 \\
Llama-2-13b-chat-hf & 4096 & 2T & 11008 & 25 GB  & 13B  & 32 \\
falcon-7b-instruct & 2048 & 1.5T & 4544 & 15 GB & 7B & 32 \\\midrule
\multicolumn{7}{c}{Ours}  \\\midrule
\model  & 256  & 4M & 384   & 604.8 MB  & 159M  & 6  \\
\bottomrule
\end{tabular}
\caption{\centering Insights about model sizes, comparing the pretrained sentence Transformers used in our approach to state-of-the-art LLMs.}
\label{tab:modelsize}
\end{table*}

Moreover, Table~\ref{tab:modelsize} shows the difference between the models we used, in terms of size, parameters, and number of layers. Our model is relatively small considering the recent advances and related work in ERC, but also compared to LLMs.

\subsection{Relative Label Representation}

Our approach actually learn twice from the data, first by using a supervised setting, and then by actually considering the relative distances between encoded element, updating through the triplet loss. This enables the use of our model to different conversation datasets with different labels. The only requirement to extend the scope of this model would be to consider another triplet sampling strategy by ignoring the labels, such as the batch-hard strategy~\cite{do2019theoretically}.

\section{Conclusion}
\label{sec:ccl}

In this paper, we present our \model model which comes from an approach mixing utterance level representation, metric learning and Siamese Networks. this model efficiently represent the conversational context, which makes it achieves state-of-the-art macroF1 score with 57.71, and satisfactory microF1 scores with 57.75 on the Emotion Recognition on DailyDialog. We also propose to use the Matthew Correlation Coefficient to better evaluate this task.

With \model we use contrastive learning with balanced samplers to overcome to minimize the imbalance factor, which is inherent to conversational data. 
We also leverage sentence bert to both minimize the memory required for training considering the whole conversational context, and to actually represent the conversational context by considering utterances as the minimal unit. This led to a more robust and efficient training method that does not require a lot of epochs to obtain satisfactory results.
We also show small to average size open source LLMs are still behind on emotion recognition in conversation as it requires a lot context to be incorporated in the prompt and is not specifically relevant to generative models.

In our future work, we want to consider applying this approach on other dataset, with added modalities in order to stress test our model. We also plan to use it on slightly different labels, as our model learns relative positions toward labels. Thus, we plan to adapt it to a more meta-learning setting.

\section*{Acknowledgments}
\label{sec:ack}

Experiments presented in this paper were carried out using the Grid’5000 testbed, supported by a scientific interest group hosted by Inria and including CNRS. RENATER and several Universities as well as other organizations (see \url{https://www.grid5000.fr}).

\section{Bibliographical References}\label{sec:reference}

\bibliographystyle{lrec-coling2024-natbib}
\bibliography{references}

\section{Language Resource References}
\label{lr:ref}
\bibliographystylelanguageresource{lrec-coling2024-natbib}
\bibliographylanguageresource{languageresource}

\end{document}